\documentclass[10pt,twocolumn,letterpaper]{article}

\usepackage[final]{cvpr}      
\definecolor{cvprblue}{rgb}{0.21,0.49,0.74}
\usepackage[pagebackref,breaklinks,colorlinks,allcolors=cvprblue]{hyperref}

\usepackage{makecell}
\usepackage{adjustbox}
\usepackage{tabularx}
\usepackage{colortbl}
\definecolor{gaingreen}{rgb}{0.0,0.50,0.25}
\definecolor{gainbg}{rgb}{0.92,0.97,0.93}
\newcommand{\gainup}[1]{\textcolor{gaingreen}{$\uparrow$#1}}
\newcommand{\gaindown}[1]{\textcolor{gaingreen}{$\downarrow$#1}}

\def\confName{CVPR}
\def\confYear{2027}

\title{GRID-UAV: Geometry and Ray-depth Integrated Decision-making for Unmanned Aerial Vehicles}

\author{%
{\normalsize Anas Anwarul Haq Khan$^{1}$ \enspace Amrin Kareem$^{1}$ \enspace Muhammad Umer Sheikh$^{1}$ \enspace Biplab Banerjee$^{2}$ \enspace Muhammad Haris Khan$^{1}$}\\[2pt]
{\small $^{1}$Mohamed bin Zayed University of Artificial Intelligence, UAE \qquad $^{2}$CSRE, IIT Bombay, India}\\[1pt]}

\newcommand{\method}{GRID}

\begin{document}
\twocolumn[{%
\renewcommand\twocolumn[1][]{#1}%
\maketitle
\begin{center}
  \includegraphics[width=\linewidth]{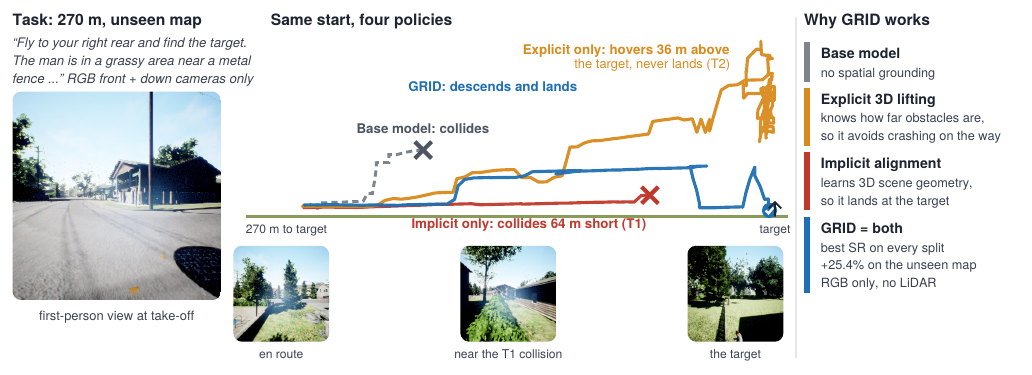}
  \captionof{figure}{One flight on an unseen map, from the same start, chosen to show the two failure. Implicit grounding alone collides on the way (T1); explicit 3D lifting alone reaches the target but hovers above it and never lands (T2); \method{} combines both and lands next to target.}
  \label{fig:teaser}
\end{center}%
}]

\begin{abstract}
Vision-language-action models (VLAs) encode images with 2D encoders that capture little 3D structure. A common approach is to feed the policy estimated depth, but a UAV chooses its own altitude, must judge distances tens of meters away, where monocular depth is less reliable, and must relate its forward and downward views. On long aerial routes, adding depth features directly brings no significant gain \textbf{(42.5\% vs.\ 42.6\% success)}, so we study how 3D information should be given to the policy. Analyzing flight failures, we find that two forms of 3D grounding fail in opposite ways: explicit 3D lifting of depth collides less on the way but more often fails to finish at the target, while implicit alignment with a 3D geometry model finishes more often but collides more. We propose \textbf{GRID}, which combines both: \textbf{(1)} it lifts monocular depth into ray-corrected 3D positions in the UAV's frame for both cameras, and \textbf{(2)} during training only, it aligns its visual features with a frozen multi-view geometry model, at no inference cost. \textbf{GRID} adding explicit lifting to implicit alignment significantly increases how often the policy reaches the target ($\textbf{61.5}\% \rightarrow \textbf{69.5}\%$, $p<0.001$), and adding implicit alignment to explicit lifting significantly increases how often it completes the task once there ($\textbf{57.7}\% \rightarrow \textbf{61.4}\%$, $p<0.05$). On TravelUAV (3,005 test episodes), GRID achieves the best success rate on every split, with relative gains over the state of the art of \textbf{9.5\% }(seen), \textbf{17.1\%} (unseen objects) and \textbf{25.4\%} (unseen map). \href{https://github.com/anas2908/GRID-UAV}{github.com/anas2908/GRID-UAV}.

\end{abstract}    
\section{Introduction}
\label{sec:intro}

Unmanned aerial vehicles (UAVs) are  increasingly used for delivery and search-and-rescue operations, where an operator would state a goal in natural language, \eg \emph{``fly to your left and find the white truck parked on a residential street,''} and let the vehicle do the rest. In aerial vision-language navigation~\cite{wang2025towards}, the UAV must follow such an instruction through a large outdoor environment, flying hundreds of meters without collision and stopping at targets it has never seen before. Because UAVs are battery-powered and payload-limited, heavy sensors such as LiDAR shorten flight time~\cite{abdilla2015power}. Thus, at test time we use only forward- and downward-facing RGB cameras and must infer 3D structure from them~\cite{sadeghi2016cad2rl,loquercio2018dronet}. VLA models~\cite{brohan2023rt,kim2024openvla,black2024pi_0,team2024octo}, now used across many robot embodiments and recently on UAVs~\cite{xu2026aerovla,sun2026autofly}, are a natural fit, but their 2D image encoders capture what is in an image rather than where it lies in 3D space~\cite{el2024probing}. This is a problem for a UAV, whose viewpoint changes rapidly in 3D, because a small tree nearby and a large tree far away can occupy the same part of the image.

A common solution is to give the policy depth information. 3D-aware VLAs~\cite{zhen20243d,qu2025spatialvla} do this successfully in robot manipulation, where the relevant objects lie close to the camera. A UAV differs in three ways. First, it chooses its own altitude, so it must reason about what lies below as well as ahead. Second, the obstacles and targets that shape its route are often tens of meters away, where monocular depth is less reliable~\cite{yin2023metric3d}. Third, near the target, the same object often appears in both cameras at once, e.g., its side in the forward view and its top in the downward one, so it must be matched across two very different views. Depth input helps UAV policies in compact, short-range settings~\cite{sun2026autofly,loquercio2021learning}, but on long outdoor flights (\cite{wang2025towards}) we find that feeding depth features directly brings no gain ($42.5\%$ w/ depth vs.\ $42.6\%$ w/o depth success), whereas lifting the depth~\cite{yang2024depthv2} into 3D~\cite{wang2019pseudo,philion2020lift} raises success to $46.6\%$. Hence, we study how 3D information should be given to UAV.

We compare two forms of 3D grounding. In \emph{explicit} grounding, we estimate metric depth with a monocular model adapted to aerial viewpoints, use each pixel's viewing ray to place every image patch at its 3D position in the UAV's own body frame, for both the forward and downward cameras, and add these positions to the visual tokens. In \emph{implicit} grounding, we align the policy's visual features during training with those of a frozen 3D geometry model~\cite{wang2025vggt}, so that it learns to represent geometry on its own. To compare the two, we look at how flights fail. A flight can fail in one of two ways: \textbf{(T1)}~the UAV never reaches the target (comes within 15\,m horizontally of it), usually because it crashes on the way; or \textbf{(T2)}~it reaches the target but fails to finish, for example by hovering high above it. The two forms fail at different stages (\cref{fig:mechanism,fig:failures}). Explicit grounding suffers less from T1: the UAV crashes less and reaches the target more often ($69.4\%$ vs.\ $61.5\%$). But it suffers more from T2: once there, it finishes less often ($57.7\%$ vs.\ $62.3\%$) and hovers above the target nearly twice as often. Implicit grounding shows the reverse. We attribute this to what each form provides. Avoiding T1 requires knowing how far away things are, which explicit grounding provides directly. Avoiding T2 requires recognizing that the object in the forward view and the one in the downward view are the same, that is, relating the two current views, which we hypothesize implicit grounding learns.

Since the two are complementary, we propose \textbf{\method{}} (\textbf{G}eometry and \textbf{R}ay-depth \textbf{I}ntegrated \textbf{D}ecision-making), a UAV VLA that combines both and needs only RGB cameras at test time. Unlike prior UAV VLAs, which either use no depth (AeroVLA~\cite{xu2026aerovla}) or feed it through a Siamese projector without 3D lifting (AutoFly~\cite{sun2026autofly}), \method{} grounds the policy in 3D both explicitly and implicitly. The explicit branch adds only $5.2$M trainable parameters (2.2\% of trainable and 0.07\% of total parameters) on top of frozen depth estimator, the implicit branch adds no inference overhead.

By combining the two (\cref{sec:analysis}), \method{} has fewer en-route failures (T1) than the implicit-only policy (reach $61.5\%\rightarrow69.5\%$, $p<0.001$) and fewer at-goal failures (T2) than the explicit-only policy (completion $57.7\%\rightarrow61.4\%$, $p<0.05$). On TravelUAV benchmark, it raises overall success from $41.7\%$ to $48.2\%$, is the best on every split, and gains most on the unseen map ($25.4\%$ relative), which tests generalization to new environments.

Our contributions are as follows:
\begin{itemize}
    \item A failure analysis of UAV VLAs that separates failing to reach the target (T1) from failing to complete the task once there (T2). We also show that directly adding depth features brings no gain in long-horizon aerial navigation, and that explicit and implicit 3D grounding address different failures.
    \item We propose \method{}, a UAV VLA policy that needs only RGB cameras at test time and combines explicit 3D lifting across forward and downward cameras with implicit alignment to a frozen geometry model, where the alignment adds no cost at inference.
    \item \method{} achieves state-of-the-art success on all TravelUAV benchmark test splits, with the largest improvement on the unseen map.
\end{itemize}

\section{Related Work}
\label{sec:related}

\paragraph{Aerial VLN benchmarks.}
Aerial VLN has been studied through dialog-guided (AVDN~\cite{fan2023aerial}), route-following (AerialVLN~\cite{liu2023aerialvln}) and map-assisted (CityNav~\cite{lee2024citynav}) benchmarks, and at scale by OpenFly~\cite{gao2026openfly} and AutoFly~\cite{sun2026autofly}. OpenFly and AutoFly use short routes (${\sim}99$\,m and ${\sim}107$\,m, the latter in $70\,\text{m}\times70\,\text{m}$ scenes); AerialVLN and CityNav have longer routes but discrete actions without flight kinematics. TravelUAV~\cite{wang2025towards}, which we use, combines long routes (${\sim}256$\,m on average, up to $1.1$\,km) with continuous flight and only a coarse direction instruction with a target description, so the UAV must itself judge distances over hundreds of meters.

\paragraph{VLAs for UAVs.}
VLAs~\cite{brohan2023rt,kim2024openvla,black2024pi_0} have been adapted to UAV racing~\cite{serpiva2025racevla}, cognitive tasks~\cite{lykov2025cognitivedrone} and navigation~\cite{sun2026autofly,xu2026aerovla}. On TravelUAV, the benchmark's own baseline pairs a navigation LLM that predicts hierarchical trajectories~\cite{wang2025towards}; NavFoM~\cite{zhang2026embodied} trains a single navigation foundation model across robot embodiments, including UAVs; and AeroVLA~\cite{xu2026aerovla}, maps forward-downward views and a coarse directional prompt directly to flight commands. All of them rely on 2D image encoders without 3D grounding.

\paragraph{3D grounding in VLAs.}
Several works add 3D information to manipulation VLAs. One line provides 3D information as input: 3D-VLA~\cite{zhen20243d} builds on 3D scene features, SpatialVLA~\cite{qu2025spatialvla} adds egocentric 3D position encodings to visual tokens, and PointVLA~\cite{li2026pointvla} and GeoVLA~\cite{sun2025geovla} inject point-cloud features. Another line transfers geometric knowledge from a 3D foundation model~\cite{wang2025vggt}, either by fusing its features into the policy (Evo-0~\cite{lin2025evo}) or by aligning the policy's internal representations~\cite{yu2025repa} with them during training (Spatial Forcing~\cite{li2026spatial}). These works target tabletop scenes close to the camera and follow one line or the other. \method{} adapts both to long-range, two-camera aerial navigation and combines them.

\paragraph{Depth for long-horizon UAV navigation.}
AutoFly~\cite{sun2026autofly} is the closest prior work: it feeds
DAv2~\cite{yang2024depthv2} to a UAV VLA through a shared (Siamese) projector
and reports a $3.9$-point gain in success, in a setting with short routes in
$70\,\text{m}\times70\,\text{m}$ scenes and a single forward camera, where
nearby obstacles keep monocular depth reliable. The TravelUAV benchmark is
more demanding: obstacles and targets lie tens to hundreds of meters away,
long routes let depth errors accumulate, and the target must be matched across two
cameras. Feeding depth through AutoFly's shared Siamese projector
(\cref{sec:depthform}) leaves success essentially unchanged
($42.6\%\rightarrow42.5\%$), whereas lifting the same depth into 3D improves
it; \method{} builds on this and adds implicit geometric alignment.

\begin{figure*}[t]
\centering
\includegraphics[width=\linewidth]{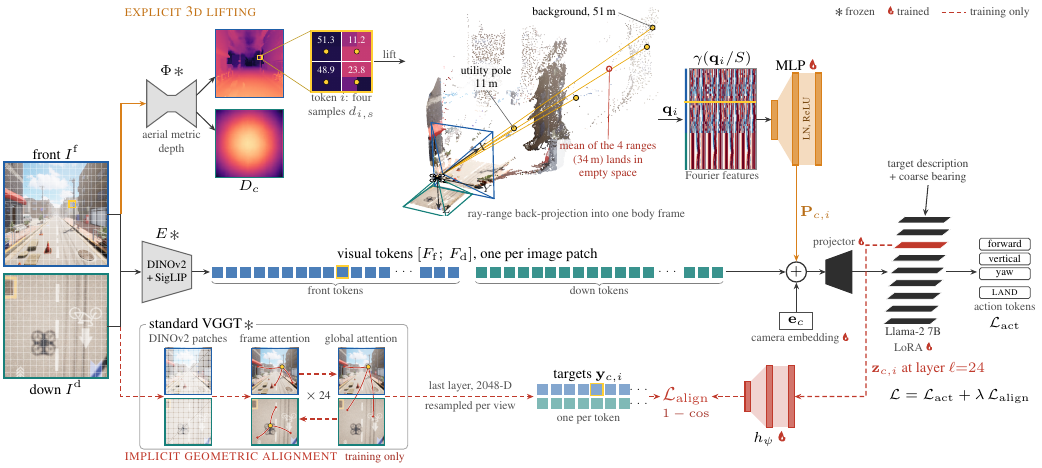}
\caption{\textbf{Overview of \method{}} on a real TravelUAV frame, following one token on a utility pole (yellow).
Top: explicit 3D lifting (\cref{sec:explicit}). The frozen depth model $\Phi$ predicts range maps $D_c$ (on the flat ground below the UAV, range grows radially).
The token keeps four depth samples $d_{i,s}$, whose mean would land in empty space. The samples are back-projected along their viewing rays into the UAV's body frame, encoded, and added to the token.
Middle: the policy (\cref{sec:base}).
Bottom: implicit alignment (\cref{sec:implicit}), used during training only (dashed). The standard VGGT aggregator~\cite{wang2025vggt}, unmodified and frozen, alternates frame attention within each view and global attention across both views 24 times (lines are schematic). Its last-layer features are resampled per view to one target $\mathbf{y}_{c,i}$ per token, and the hidden states $\mathbf{z}_{c,i}$ at LLM layer $\ell{=}24$ are aligned with these targets.}
\label{fig:method}
\end{figure*}

\section{Methodology}
\label{sec:method}

\method{} keeps a standard VLA policy and changes only what its visual tokens carry (\cref{fig:method}).
\emph{Explicit 3D lifting} (\cref{sec:explicit}) adds to each token the metric position of the surface it sees, and \emph{implicit geometric alignment} (\cref{sec:implicit}) aligns the same tokens inside the LLM with a frozen VGGT during training only.
Both need only the RGB cameras at test time and leave the action interface unchanged.

\subsection{Task and Base Policy}
\label{sec:base}

\paragraph{Task.}
At each step, the UAV observes a front image $I^{\mathrm{f}}$ and a downward image $I^{\mathrm{d}}$.
It is given a description of the target and, as in~\cite{xu2026aerovla}, a coarse bearing toward it (one of seven sectors, e.g.\ ``forward-left''), recomputed at every step from the target's true position, which makes it a coarse form of oracle guidance.
The policy outputs forward, vertical and yaw commands, or \texttt{LAND} to end the episode; an episode succeeds if the UAV stops within 20\,m of the target~\cite{wang2025towards}.

\paragraph{Base policy.}
We follow the formulation in~\cite{xu2026aerovla} on an OpenVLA backbone~\cite{kim2024openvla}.
The frozen vision encoder $E$ (DINOv2~\cite{oquab2023dinov2} and SigLIP~\cite{zhai2023sigmoid}) encodes each view $c\in\{\mathrm{f},\mathrm{d}\}$ separately, with shared weights, into $N{=}16{\times}16$ patch tokens~\cite{dosovitskiy2020image} $F_c\in\mathbb{R}^{N\times C}$, so every token covers part of exactly one camera.
The concatenated tokens $[F_{\mathrm{f}};F_{\mathrm{d}}]$ pass through a trainable projector into Llama-2 7B~\cite{touvron2023llama}, fine-tuned with LoRA~\cite{hu2021lora}.
Each command is quantized into 99 uniform bins and written as a two-digit token, and training minimizes the cross-entropy $\mathcal{L}_{\mathrm{act}}$ of these action tokens.

\subsection{Explicit 3D Lifting}
\label{sec:explicit}

We add to every token the metric position of the surface it sees, in the UAV's own frame (e.g., the utility pole in \cref{fig:method}).
The idea is inspired by egocentric position encodings for tabletop manipulation~\cite{qu2025spatialvla}, but three properties of flight shape the steps below: wide-angle cameras, several views that must share one frame, and a wide range of depths within one view.
We describe one token $i$ of view $c$ (in \cref{fig:method}, the highlighted token on the pole) and omit $c$ where it is clear.

\paragraph{Aerial metric depth.}
Metric Depth Anything V2-Large~\cite{yang2024depthv2} is trained on street-level driving scenes, far from aerial views.
We fine-tune its encoder and head with the SiLog loss~\cite{eigen2014depth} on simulator depth~\cite{shah2017airsim} of front and downward images from the training split only, one view at a time as at test time, and freeze it as $\Phi$.
Since $\Phi$ learns from the simulator's depth, which is the range along each pixel's viewing ray, every value of $D_c{=}\Phi(I^c)$ is a metric range.
At test time, depth comes from RGB alone, with no depth sensor or ground truth; \cref{sec:analysis} compares $\Phi$ with the off-the-shelf model.

\paragraph{Four samples per token.}
As in ~\cite{qu2025spatialvla}, each token keeps four separate depth samples instead of a single depth.
The map $D_c$ is resized to $64{\times}64$, so each token covers a $4{\times}4$ tile, which we split into $2{\times}2$ sub-tiles $T_{i,s}$.
Each sub-tile gives one sample, $d_{i,s}=\operatorname{mean}_{\mathbf{p}\in T_{i,s}}D_c(\mathbf{p})$ for $s=1,\dots,4$, located at its center $\mathbf{p}_{i,s}=(u_{i,s},v_{i,s})$ in original image coordinates.
We never average the four samples into one depth: each is lifted to its own 3D point.
This is important in flight, where depth varies widely within a token: the pole token's samples span 11.2\,m on the pole to 51.3\,m on the background, and a single depth for the token (their mean, 34\,m) would place it in empty space (red circle in \cref{fig:method}).

\paragraph{Ray-range back-projection into one body frame.}
Each sample is lifted along its unit viewing ray and moved into a body frame shared by both cameras:
\begin{equation}
  \mathbf{x}_{i,s} = R_c\, d_{i,s}\,\frac{K^{-1}\tilde{\mathbf{p}}_{i,s}}{\lVert K^{-1}\tilde{\mathbf{p}}_{i,s}\rVert_2} + \mathbf{t}_c ,
  \label{eq:backproject}
\end{equation}
where $\tilde{\mathbf{p}}{=}[u,v,1]^\top$ is the homogeneous pixel, $K$ holds the pinhole intrinsics~\cite{hartley2003multiple}, and $(R_c,\mathbf{t}_c)$ is the fixed mounting of camera $c$.
The ray is normalized because $d_{i,s}$ is a range, not a depth along the optical axis; the usual form $d\,K^{-1}\tilde{\mathbf{p}}$ would stretch points by up to $1.7\times$ toward the corners of our $90^\circ$ cameras.
The body frame moves with the UAV ($X$ forward, $Y$ right, $Z$ down), so a surface gets the same coordinates from either camera without any map or global pose, and further calibrated cameras can be added the same way.
The four points form the token's descriptor $\mathbf{q}_i=[\mathbf{x}_{i,1}^\top,\dots,\mathbf{x}_{i,4}^\top]^\top\in\mathbb{R}^{12}$.

\paragraph{Fourier encoding and fusion.}
A small MLP struggles to represent fine and coarse structure at once from raw coordinates~\cite{tancik2020fourier}, so each coordinate is first expanded into Fourier features:
\begin{align}
  \gamma(z) &= \big[z,\ \sin(2^k z),\ \cos(2^k z)\big]_{k=0}^{L-1}, \nonumber\\
  \mathbf{P}_{c,i} &= \mathrm{MLP}\big(\gamma(\mathbf{q}_i/S)\big), \label{eq:fourier}\\
  \hat{F}_{c,i} &= F_{c,i} + \mathbf{P}_{c,i} + \mathbf{e}_c . \label{eq:fuse}
\end{align}
With $L{=}8$ and $S{=}40$\,m, the periods $2\pi S/2^k$ span 2 to 250\,m, from metre-scale clearance to the farthest predicted depths.
The MLP (Linear--LayerNorm--ReLU--Linear) maps the code to the token width $C$, $\mathbf{e}_c$ is a learned camera embedding, and the lifted tokens $[\hat{F}_{\mathrm{f}};\hat{F}_{\mathrm{d}}]$ replace $[F_{\mathrm{f}};F_{\mathrm{d}}]$ at the projector input.
Since the last MLP layer and $\mathbf{e}_c$ are zero-initialized, training starts from the base policy.
The branch leaves the token count and the LLM unchanged and adds 5.2M parameters (2.2\% of the trainable and 0.07\% of all parameters) plus one frozen depth pass per view.

\subsection{Implicit Geometric Alignment}
\label{sec:implicit}

Each token is lifted independently, so the policy must still learn from $\mathcal{L}_{\mathrm{act}}$ alone which tokens show the same object.
This is hardest near the goal, where the policy decides whether to land and the target is often split between the cameras, \eg a car's side at the bottom of the front image and its roof in the downward image.
VGGT~\cite{wang2025vggt} computes each view's features in the context of
the other through global attention, so its targets carry both per-view geometry
and cross-view context; we use it as a frozen teacher~\cite{hinton2015distilling} during training.

\paragraph{Standard VGGT teacher.}
VGGT is used as released (\cref{fig:method}, bottom): both views enter it together at 518\,px, DINOv2 splits each into $37{\times}37$ patch tokens, and 24 rounds of frame attention, within a view, and global attention, across views, compute each view's features in the context of the other (in \cref{fig:method}, global attention links the pole token to the downward image).
Unlike $\Phi$, VGGT never feeds the policy, so we do not fine-tune it.

\paragraph{Targets and loss.}
Each token gets one VGGT target, which the policy learns to match:
\begin{align}
  \mathbf{y}_c &= \mathcal{B}(G_c), \quad [G_{\mathrm{f}};G_{\mathrm{d}}] = \mathrm{VGGT}(I^{\mathrm{f}},I^{\mathrm{d}}), \label{eq:target}\\
  \mathcal{L}_{\mathrm{align}} &= 1-\frac{1}{2N}\sum_{c,i}\cos\big(h_\psi(\mathbf{z}_{c,i}),\,\mathbf{y}_{c,i}\big), \label{eq:align}\\
  \mathcal{L} &= \mathcal{L}_{\mathrm{act}}+\lambda\,\mathcal{L}_{\mathrm{align}} . \label{eq:total}
\end{align}
Here $G_c\in\mathbb{R}^{37\times37\times2048}$ holds view $c$'s patch features from the last aggregator layer, plus VGGT's 2D position embedding scaled by 0.1, as in~\cite{li2026spatial}.
The bilinear resampling $\mathcal{B}$ maps each view separately to the policy's $16{\times}16$ grid, so $\mathbf{y}_{c,i}$ covers the same image area as token $(c,i)$ (highlighted for the pole token).
On the policy side, $\mathbf{z}_{c,i}$ is the hidden state of token $(c,i)$ at layer $\ell{=}24$ of the 32-layer LLM, and the head $h_\psi$ (LayerNorm--Linear--GELU--Linear) maps it to 2048-D; since the comparison goes through $h_\psi$ and ignores scale, the hidden states need not equal the VGGT features.
Following that, we align a deep but not final layer and set $\lambda{=}0.5$; geometry imposed on shallow layers fades higher up, and the last layers keep little visual content.
In the causal LLM~\cite{vaswani2017attention}, the downward tokens follow the front tokens and attend to them, which lets them absorb the cross-view context of their targets.
The gradient also reaches the earlier layers, the projector and the explicit branch, so both components are trained jointly.

\paragraph{Inference.}
VGGT and $h_\psi$ (12.6M parameters) are discarded after training, so the deployed policy is the model of \cref{sec:explicit} with no added parameters or computation.

\providecommand{\std}[1]{{\scriptsize$\,\pm$#1}}
\providecommand{\aerore}{AeroVLA$^{\ddagger}$}

\section{Experiments}
\label{sec:experiments}
\label{sec:analysis}

We evaluate \method{} on TravelUAV: we compare it with the state of the art (\cref{sec:main}), study how depth should be integrated into the policy (\cref{sec:depthform}), compare explicit and implicit grounding and analyze their failures (\cref{sec:expimp,sec:failures}), and report the computational cost of each component (\cref{sec:cost}).

\subsection{Experimental Setup}
\label{sec:setup}

\paragraph{Benchmark.}
We use TravelUAV's UAV-Need-Help task~\cite{wang2025towards} with AeroVLA's curated training set~\cite{xu2026aerovla} (7,611 trajectories, 410k frames, 11 environments).
The 3,005 test episodes form Seen (1,418), Unseen Object (629) and Unseen Map (958) splits, each divided into Easy ($\leq$250\,m) and Hard ($>$250\,m) by reference path length; routes average 256\,m and reach 1.1\,km.

\paragraph{Metrics.}
We report the benchmark's metrics~\cite{wang2025towards}: success rate (SR; the episode ends within 20\,m of the target), oracle SR (OSR; within 20\,m at any time), navigation error (NE; final distance in meters) and success weighted by path length, $\mathrm{SPL}=\frac{1}{N}\sum_i S_i\,\ell_i/\max(p_i,\ell_i)$, with reference length $\ell_i$ and flown length $p_i$.

\paragraph{Baselines.}
We compare with CMA~\cite{anderson2018vision}, the TravelUAV model~\cite{wang2025towards} and NavFoM~\cite{zhang2026embodied}, which use the benchmark's dense oracle guidance, a more detailed form than the coarse bearing given to \method{} (numbers from~\cite{xu2026aerovla}).
All of these baselines were evaluated on headless GPUs, \ie, without an attached display, as is standard in data centers.
The AeroVLA report lower success rates in this setting than with a physical display, which authors recommend, because off-screen rendering drops frames and destabilizes the physics.
Display GPUs are rarely available on research clusters, so for a fair comparison with the other baselines we evaluate every method headless, using the checkpoint released by the AeroVLA authors \textit{(Supplementary, Section C)}.
AutoFly-depth follows AutoFly~\cite{sun2026autofly}: aerial-adapted DAv2 depth enters through the projector shared with the RGB features, without 3D lifting.

\paragraph{Implementation and statistical testing.}
All our policies start from OpenVLA-7B~\cite{kim2024openvla} and are trained with identical settings on four A100 (40\,GB) GPUs; \cref{tab:hparams} lists all hyperparameters.
All methods are evaluated on the same 3,005 episodes from identical start states, so we compare them with paired tests: an exact McNemar test~\cite{mcnemar1947note} on per-episode success and 95\% confidence intervals from a paired bootstrap over episodes (10,000 resamples).
We repeat every ablation evaluation three times and report the standard deviation across runs (\cref{tab:ablation}), together with the largest $p$-value over runs.
Completion is measured only on the episodes in which the UAV reaches the target, about 590--670 per model on the unseen map. Repeated evaluation runs of the same policy also differ (\Cref{tab:ablation}), and on fewer episodes this noise matters more. For quantities conditioned on reaching the target, we therefore pool five evaluation runs per model on the unseen map and resample whole episodes in the bootstrap, so that the extra runs reduce evaluation noise without counting as additional episodes.

\begin{table*}[t]
\centering
\footnotesize
\setlength{\tabcolsep}{3.4pt}
\caption{\textbf{Comparison with prior work on TravelUAV.}
SR, OSR and SPL in \%, NE in meters.
AeroVLA results are reproduced from the released checkpoint.
Bold: best in each column.
Rel.\ gain: relative improvement of \method{} over SoTA.
The gain of \method{} over SoTA is significant on every split and subset ($\{$Seen, Unseen Object, Unseen Map$\}$ $\times$ $\{$Full, Easy, Hard$\}$; exact McNemar test, $p{<}0.05$) except Test Seen Hard.}
\label{tab:main}
\begin{adjustbox}{max width=\linewidth}
\begin{tabular}{@{}l cccc cccc cccc@{}}
\toprule
& \multicolumn{4}{c}{Full} & \multicolumn{4}{c}{Easy} & \multicolumn{4}{c}{Hard} \\
\cmidrule(lr){2-5}\cmidrule(lr){6-9}\cmidrule(l){10-13}
Method & NE$\downarrow$ & SR$\uparrow$ & OSR$\uparrow$ & SPL$\uparrow$ & NE$\downarrow$ & SR$\uparrow$ & OSR$\uparrow$ & SPL$\uparrow$ & NE$\downarrow$ & SR$\uparrow$ & OSR$\uparrow$ & SPL$\uparrow$ \\
\midrule
\multicolumn{13}{@{}l}{\textit{Test Seen} (1,418 episodes)} \\
CMA (CVPR'18)~\cite{anderson2018vision} & 135.73 & 8.37 & 18.72 & 7.90 & 84.89 & 11.48 & 24.52 & 10.68 & 197.77 & 4.57 & 11.65 & 4.51 \\
TravelUAV (ICLR'25)~\cite{wang2025towards} & 98.66 & 17.45 & 48.87 & 15.76 & 66.40 & 20.26 & 51.23 & 18.10 & 138.04 & 14.02 & 45.98 & 12.90 \\
NavFoM (ICLR'26)~\cite{zhang2026embodied} & 93.05 & 29.17 & 49.24 & 25.03 & 58.98 & 32.91 & 53.16 & 27.87 & 143.83 & 23.58 & 43.40 & 20.80 \\
AeroVLA (ECCV'26)~\cite{xu2026aerovla} & \underline{75.63} & \underline{43.86} & \underline{52.82} & \underline{34.37} & \underline{47.58} & \underline{45.34} & \underline{56.58} & \underline{33.34} & \underline{110.22} & \underline{42.05} & \underline{48.19} & \underline{35.64} \\
\textbf{\method{} (ours)} & \textbf{68.83} & \textbf{48.03} & \textbf{56.35} & \textbf{38.15} & \textbf{44.32} & \textbf{50.32} & \textbf{60.66} & \textbf{37.55} & \textbf{99.05} & \textbf{45.20} & \textbf{51.02} & \textbf{38.89} \\
\rowcolor{gainbg}
\textcolor{gaingreen}{\textit{Rel.\ gain (\%)}} & \gaindown{9.0} & \gainup{9.5} & \gainup{6.7} & \gainup{11.0} & \gaindown{6.9} & \gainup{11.0} & \gainup{7.2} & \gainup{12.6} & \gaindown{10.1} & \gainup{7.5} & \gainup{5.9} & \gainup{9.1} \\
\midrule
\multicolumn{13}{@{}l}{\textit{Test Unseen Object} (629 episodes)} \\
CMA (CVPR'18)~\cite{anderson2018vision} & 155.79 & 9.06 & 16.06 & 8.68 & 102.92 & 14.83 & 22.49 & 13.90 & 182.09 & 6.19 & 12.86 & 6.08 \\
TravelUAV (ICLR'25)~\cite{wang2025towards} & 118.11 & 22.42 & 46.90 & 20.51 & 86.12 & 24.40 & 49.28 & 22.03 & 134.03 & 21.43 & 45.71 & 19.75 \\
NavFoM (ICLR'26)~\cite{zhang2026embodied} & 108.04 & 29.83 & 47.99 & 27.20 & 70.51 & 32.54 & 50.72 & 29.54 & 133.01 & 28.03 & 46.18 & 25.64 \\
AeroVLA (ECCV'26)~\cite{xu2026aerovla} & \underline{73.83} & \underline{48.49} & \underline{55.80} & \underline{39.06} & \underline{54.16} & \underline{45.93} & \underline{54.55} & \underline{35.46} & \underline{83.62} & \underline{49.76} & \underline{56.43} & \underline{40.85} \\
\textbf{\method{} (ours)} & \textbf{68.32} & \textbf{56.76} & \textbf{61.37} & \textbf{46.71} & \textbf{48.30} & \textbf{58.85} & \textbf{64.11} & \textbf{45.03} & \textbf{78.28} & \textbf{55.71} & \textbf{60.00} & \textbf{47.55} \\
\rowcolor{gainbg}
\textcolor{gaingreen}{\textit{Rel.\ gain (\%)}} & \gaindown{7.5} & \gainup{17.1} & \gainup{10.0} & \gainup{19.6} & \gaindown{10.8} & \gainup{28.1} & \gainup{17.5} & \gainup{27.0} & \gaindown{6.4} & \gainup{12.0} & \gainup{6.3} & \gainup{16.4} \\
\midrule
\multicolumn{13}{@{}l}{\textit{Test Unseen Map} (958 episodes)} \\
CMA (CVPR'18)~\cite{anderson2018vision} & 141.68 & 2.30 & 10.02 & 2.16 & 102.29 & 3.57 & 14.26 & 3.33 & 197.35 & 0.50 & 4.03 & 0.50 \\
TravelUAV (ICLR'25)~\cite{wang2025towards} & 138.80 & 4.18 & 20.77 & 3.84 & 102.94 & 4.63 & 22.82 & 4.24 & 189.46 & 3.53 & 17.88 & 3.28 \\
NavFoM (ICLR'26)~\cite{zhang2026embodied} & 125.10 & 6.30 & 18.95 & 5.68 & 102.41 & 6.77 & 20.07 & 6.04 & 170.58 & 5.36 & 15.71 & 4.97 \\
AeroVLA (ECCV'26)~\cite{xu2026aerovla} & \underline{70.86} & \underline{34.03} & \underline{52.30} & \underline{24.37} & \underline{50.51} & \underline{39.22} & \underline{57.58} & \underline{26.68} & \underline{99.63} & \underline{26.70} & \underline{44.84} & \underline{21.11} \\
\textbf{\method{} (ours)} & \textbf{58.14} & \textbf{42.69} & \textbf{59.39} & \textbf{31.77} & \textbf{44.89} & \textbf{44.74} & \textbf{61.68} & \textbf{31.20} & \textbf{76.85} & \textbf{39.80} & \textbf{56.17} & \textbf{32.57} \\
\rowcolor{gainbg}
\textcolor{gaingreen}{\textit{Rel.\ gain (\%)}} & \gaindown{18.0} & \gainup{25.4} & \gainup{13.6} & \gainup{30.4} & \gaindown{11.1} & \gainup{14.1} & \gainup{7.1} & \gainup{16.9} & \gaindown{22.9} & \gainup{49.1} & \gainup{25.3} & \gainup{54.3} \\
\bottomrule
\end{tabular}
\end{adjustbox}
\end{table*}

\subsection{Comparison with the State of the Art}
\label{sec:main}

\Cref{tab:main} compares \method{} with prior work.
Under identical evaluation settings, \method{} improves SR over AeroVLA significantly on every split, from 41.7\% to 48.2\% over all episodes ($p{=}3{\times}10^{-9}$), together with NE, OSR and SPL.
The gain grows with novelty, from Seen to the unseen map (25.4\% relative), and peaks on its long Hard routes (26.7\% to 39.8\%, $p{<}10^{-4}$); baselines with dense oracle guidance remain far behind.
Although $\Phi$ is fine-tuned on simulator depth (\cref{sec:explicit}), the best variants without it still reach 45.8\% SR (\Cref{tab:ablation}).



\subsection{Depth Integration}
\label{sec:depthform}

\begin{figure}[t]
\centering
\includegraphics[width=\linewidth]{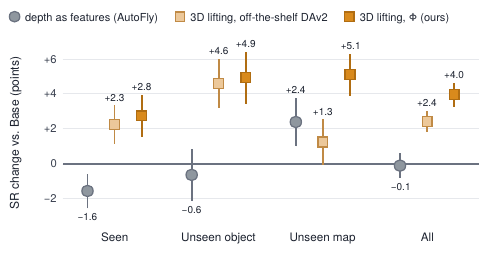}
\caption{\textbf{Effect of depth integration.} SR change over Base (no depth), with the standard deviation over 3 evaluation runs. Depth  (AutoFly's siamese projector) brings no gain overall; lifting depth helps on every split, most with $\Phi$ on the unseen map.}

\label{fig:depthform}
\end{figure}

\Cref{tab:ablation}(a) and \cref{fig:depthform} test how depth should be given to the policy.
All variants start from the same policy without depth (Base), which sees the front and down views separately.
Base + AutoFly's depth feeds depth through a Siamese projector, as in AutoFly~\cite{sun2026autofly}.
The explicit variants instead use depth to place each token in 3D.
\textit{Supplementary, Section B.1} reports all metrics for every variant on the Full, Easy and Hard episodes of each split.

\paragraph{Depth as input features.}
Base + AutoFly's depth reaches 42.5\% SR against 42.6\% for Base ($p{=}0.95$), with no significant difference on any split except the Hard routes of the unseen map.
This holds even though its depth comes from our aerial-adapted $\Phi$, which is far more accurate than off-the-shelf DAv2, so the lack of gain is not due to poor depth.
AutoFly reports a 3.9-point gain from depth on routes of 107\,m on average~\cite{sun2026autofly}; on TravelUAV's routes, which are more than twice as long, appending depth to the visual input does not help the policy.

\begin{table*}[t]
\centering
\footnotesize
\setlength{\tabcolsep}{4pt}
\caption{\textbf{Ablations.} SR (\%) per split and over all episodes, with the standard deviation over three evaluation runs, and SPL over all episodes.
Views: one mosaic image or separate views. Depth: none, depth features, or depth lifted into 3D.
$\Phi$: aerial-adapted depth model (otherwise off-the-shelf DAv2). Impl.: implicit alignment.
$p$: exact McNemar test of SR against \method{} over all episodes (largest over runs).}
\label{tab:ablation}
\begin{adjustbox}{max width=\linewidth}
\begin{tabular}{@{}l cccc cccc c c@{}}
\toprule
Method & Views & Depth & $\Phi$ & Impl. & Seen & Unseen Obj. & Unseen Map & All & SPL & $p$ \\
\midrule
\multicolumn{11}{@{}l}{\textit{(a) Depth integration}} \\
AeroVLA (ECCV'26) & mosaic & -- & -- & -- & 43.86\std{1.1} & 48.49\std{1.6} & 34.03\std{1.3} & 41.70\std{0.7} & 32.16 & $3{\times}10^{-9}$ \\
Base (no depth) & sep. & -- & -- & -- & 44.71\std{1.2} & 49.60\std{1.7} & 34.97\std{1.3} & 42.63\std{0.8} & 32.87 & $4{\times}10^{-7}$ \\
Base + AutoFly's depth (ICLR'26) & sep. & feat. & \checkmark & -- & 43.16\std{1.0} & 48.97\std{1.5} & 37.37\std{1.4} & 42.53\std{0.7} & 33.04 & $2{\times}10^{-7}$ \\
Explicit, off-the-shelf depth & sep. & 3D & -- & -- & 46.97\std{1.1} & 54.21\std{1.4} & 36.22\std{1.3} & 45.06\std{0.6} & 36.69 & 0.007 \\
Explicit & sep. & 3D & \checkmark & -- & \underline{47.46}\std{1.2} & 54.53\std{1.5} & \underline{40.08}\std{1.2} & \underline{46.59}\std{0.7} & 36.65 & 0.15 \\
\midrule
\multicolumn{11}{@{}l}{\textit{(b) Adding implicit alignment}} \\
Implicit & sep. & -- & -- & \checkmark & 47.10\std{1.0} & 54.37\std{1.6} & 38.31\std{1.4} & 45.82\std{0.7} & \underline{37.57} & 0.03 \\
Explicit (off-the-shelf) + implicit & sep. & 3D & -- & \checkmark & 47.39\std{1.1} & \underline{54.69}\std{1.5} & 37.47\std{1.3} & 45.76\std{0.7} & 36.91 & 0.04 \\
\textbf{\method{}} (explicit + implicit) & sep. & 3D & \checkmark & \checkmark & \textbf{48.03}\std{0.9} & \textbf{56.76}\std{1.4} & \textbf{42.69}\std{1.2} & \textbf{48.15}\std{0.6} & \textbf{37.91} & -- \\
\bottomrule
\end{tabular}
\end{adjustbox}
\end{table*}

\paragraph{Depth lifted into 3D.}
Lifting depth into egocentric 3D positions improves SR over Base by 5.7\% with off-the-shelf DAv2 and by 9.3\% with $\Phi$ (relative).
Off-the-shelf DAv2 misjudges metric range badly from the air \textit{(Supplementary, Section A)}, twenty times worse than $\Phi$, yet still helps, which suggests that the encoding MLP (\cref{eq:fourier}) learns to correct a consistent scale error.
The benefit of accurate depth is largest on the unseen map, where $\Phi$ improves SR over off-the-shelf DAv2 by 10.7\% (relative).



\subsection{Explicit and Implicit Grounding}
\label{sec:expimp}

\Cref{tab:ablation}(b) compares the two forms of grounding and their combination.
Both improve SR over Base: the implicit-only model by 7.5\% and the explicit-only model by 9.3\% (relative).
\method{}, which uses both, improves SR by 3.3\% over the explicit-only model and by 5.1\% over the implicit-only model, and is the best model on every split.
With off-the-shelf depth, adding implicit alignment to the explicit-only model likewise improves SR, by 1.6\%.
Over all episodes, however, the SR gain of \method{} over the explicit-only model is not significant ($p=0.15$), and we do not claim that combining the two forms significantly improves overall SR over the stronger one.
Our claim is instead that the two forms fail at different stages of a flight and that \method{} significantly reduces the main failure of each, which we test on the unseen map below: \method{} reaches the target more often than the implicit-only model (fewer T1 failures, $p<0.001$) and, once there, completes the task more often than the explicit-only model (fewer T2 failures, $p<0.05$).
Because \method{} and the explicit-only model reach the target equally often, this second gain affects only the episodes that reach the target, which limits its effect on overall SR.

\begin{figure}[t]
\centering
\includegraphics[width=\linewidth]{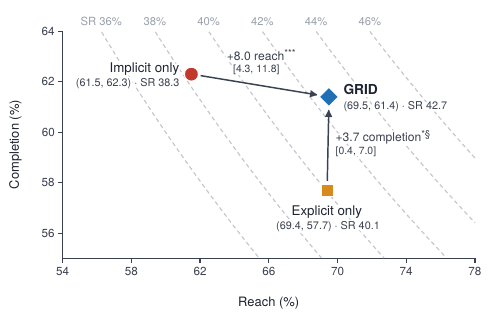}
\caption{\textbf{Reach and completion} on the unseen map. Reach: within 15\,m (horizontally) of the target; completion: reached episodes that succeed. Points: (reach, completion) and SR, in \%; dashed curves: equal $\mathrm{SR}=\text{reach}\times\text{completion}$. Arrows: paired gains of \method{}, with 95\% bootstrap CIs in brackets ($^{***}p{<}0.001$, exact McNemar test; $^{*}p{<}0.05$, paired bootstrap; $^\S$five runs).}

\label{fig:mechanism}
\end{figure}


\subsection{Failure Analysis}
\label{sec:failures}

\paragraph{Decomposition.}
We split each episode at the first moment the UAV comes within 15\,m (horizontally) of the target; the results also hold for 10\,m and 20\,m \textit{(Supplementary, Section B.2)}.
\emph{Reach} is the fraction of episodes that get there, \emph{completion} the fraction of these that succeed, and $\mathrm{SR}=\text{reach}\times\text{completion}$.
A failure is T1 if the target is never reached (collision, or lost: drifting away, stuck or out of steps) and T2 otherwise (hovering within 15\,m horizontally but more than 20\,m above the target, stopping in the wrong place after having been within the 20\,m success radius, or other).
Labels are assigned automatically from the flight logs and confirmed by manually reviewing the replays of all 1{,}714 failed unseen-map episodes of the implicit-only, explicit-only and \method{} models (about 60 hours).
We analyze the unseen map, where train-test environment overlap does not confound the effect of either form of grounding: in seen and unseen-object episodes, whose environments appear in training, a policy can partly rely on familiar layouts instead of inferring 3D structure, whereas on unseen map it judges distances and positions from  current images alone.

\begin{figure}[t]
\centering
\includegraphics[width=\linewidth]{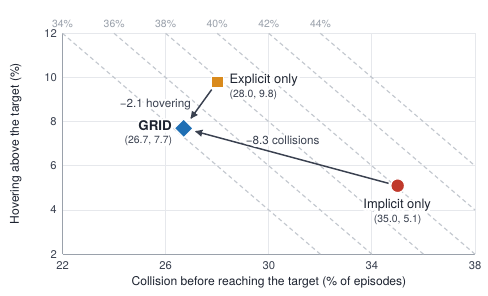}
\caption{\textbf{Failure modes} on the unseen map (\% of 958 episodes). Points give (collision, hovering); arrows give the change to \method{} in points; dashed lines mark an equal sum of the two rates, and lower left is better. The other failure types differ little between the models (stopping in the wrong place 16.3--17.1\%, lost 2.6--3.8\%, other 1.3--2.8\%).}

\label{fig:failures}
\end{figure}

\paragraph{Failures of explicit and implicit grounding.}
Explicit lifting reaches the target more often than implicit alignment (69.4\% vs.\ 61.5\%; $+7.9$ points, 95\% CI $[4.3, 11.6]$; \cref{fig:mechanism}), mainly because it collides less on the way (28.0\% vs.\ 35.0\% of episodes, \cref{fig:failures}), but once there it completes the task less often (57.7\% vs.\ 62.3\%; $-4.6$, $[-8.0, -1.2]$) and hovers high above the target nearly twice as often (9.8\% vs.\ 5.1\%). The two effects largely cancel, so the SRs of the two models differ by only 1.8 points.
This is consistent with what each form provides: explicit lifting gives distances to obstacles, which helps avoid T1 failures, and we hypothesize that implicit alignment, whose teacher computes geometry over both views jointly, helps avoid T2 failures.

\paragraph{Combining both.}
\method{} reaches the target as often as the explicit model ($+0.1$, $[-3.8, 3.9]$) and 8.0 points more often than the implicit model ($[4.3, 11.8]$, $p{<}0.001$), as collisions fall from 35.0\% to 26.7\% of episodes ($p{<}10^{-4}$).
It completes the task nearly as often as the implicit model ($-0.9$, $[-5.8, 4.0]$) and 3.7 points more often than the explicit model ($p{<}0.05$ over five runs), as hovering drops from 9.8\% to 7.7\%.

\paragraph{Remaining failures.}
Stopping in the wrong place affects 16--17\% of episodes for all three models; it is the second most common failure after collisions and the largest one that neither form of grounding reduces.
\method{} also gets lost slightly more often than the explicit model (3.8\% vs.\ 2.6\%), a small effect that involves few episodes.

\begin{table}[t]
\centering
\footnotesize
\setlength{\tabcolsep}{3pt}
\caption{\textbf{Computational cost.} Trainable parameters ($^{\ast}$used during training only); training on four A100 (40\,GB) GPUs for 32,065 steps; median compute time per decision (pre-processing, including depth, and policy forward) and peak memory on one A100, measured over 120 episodes. All models encode the two views separately (512 visual tokens).}
\label{tab:cost}
\begin{adjustbox}{max width=\linewidth}
\begin{tabular}{@{}l cccc@{}}
\toprule
Model & Trainable & GPU-hours & ms / decision & VRAM (GiB) \\
\midrule
Base & 231.3M & 171 & 522 & 15.1 \\
Explicit & 236.5M & 190 & 674 & 16.4 \\
Implicit & 231.3M\,+\,12.6M$^{\ast}$ & 242 & 522 & 15.1 \\
\textbf{\method{}} & 236.5M\,+\,12.6M$^{\ast}$ & 242 & 674 & 16.4 \\
\bottomrule
\end{tabular}
\end{adjustbox}
\end{table}

\begin{table}[t]
\centering
\footnotesize
\setlength{\tabcolsep}{5pt}
\caption{\textbf{Hyperparameters.} Identical for all policies.}
\label{tab:hparams}
\begin{adjustbox}{max width=\linewidth}
\begin{tabular}{@{}ll@{}}
\toprule
\multicolumn{2}{@{}l}{\textit{Policy}} \\
Initialization & OpenVLA-7B~\cite{kim2024openvla} \\
Frozen & vision encoders \\
Trained & projector, explicit branch, LoRA \\
LoRA rank, $\alpha$, dropout & 64, 128, 0.05 (q, k, v, o, gate, up, down) \\
Optimizer & AdamW, lr $2{\times}10^{-4}$, weight decay 0.03 \\
Schedule & cosine, 3\% warm-up, 5 epochs (32,065 steps) \\
Global batch & 64 (4 GPUs $\times$ 1 $\times$ 16 accumulation) \\
Precision, seed & bfloat16, 42 \\
\midrule
\multicolumn{2}{@{}l}{\textit{Explicit lifting}} \\
Samples per token & 4 ($2{\times}2$ sub-tiles of a $4{\times}4$ depth tile) \\
Fourier bands $L$, scale $S$ & 8, 40\,m \\
Final MLP layer, camera embedding & zero-initialized \\
\midrule
\multicolumn{2}{@{}l}{\textit{Implicit alignment (training only)}} \\
Teacher & VGGT-1B, last aggregator layer, frozen \\
Student layer $\ell$, weight $\lambda$ & 24 of 32, 0.5 \\
Head & LayerNorm--Linear--GELU--Linear \\
\midrule
\multicolumn{2}{@{}l}{\textit{Aerial depth model $\Phi$}} \\
Initialization & DAv2-L, metric (Virtual KITTI) \\
Loss, schedule & SiLog, 3 epochs (61,329 steps) \\
Batch, lr (head / encoder) & 16 front/down pairs, $5{\times}10^{-5}$ / $5{\times}10^{-6}$ \\
\bottomrule
\end{tabular}
\end{adjustbox}
\end{table}

\subsection{Computational Cost}
\label{sec:cost}

\Cref{tab:cost} compares training and inference costs with Base (no grounding).
The explicit branch adds 5.2M trainable parameters and one frozen depth pass (335M parameters) per view, costing about 150\,ms per decision; implicit alignment adds a 12.6M-parameter head and a frozen VGGT pass during training only, raising training cost but not inference cost.
Fine-tuning $\Phi$ is a one-time cost of 48 GPU-hours (training settings in \cref{tab:hparams}).

\section{Limitations and Future Work}
\label{sec:limitations}

All experiments run in simulation; adapting \method{} to real sensor noise, wind and lighting changes is the next step.
Like prior UAV VLAs, we evaluate only on TravelUAV, since short-route benchmarks (\eg, about 100\,m in AutoFly~\cite{sun2026autofly}) would not test the long-route failures we study; further long-horizon benchmarks are future work.
Each policy is trained once, so repeated evaluations and paired tests capture evaluation noise but not training variance, which would require retraining a 7B policy (up to 242 GPU-hours each; \cref{tab:cost}).
Stopping in the wrong place remains an error source (\cref{sec:failures}).
Like every VLA-based baseline, \method{} inherits a large latency (674\,ms per decision).

\section{Conclusion}
\label{sec:conclusion}

We studied how a UAV VLA should receive 3D information.
On long aerial routes, appending depth features to the visual input brings no gain, whereas lifting depth into 3D does.
Separating failures to reach the target (T1) from failures to complete the task (T2) showed that explicit 3D lifting mainly reduces the first and implicit alignment the second.
\method{} combines both and, using only RGB cameras at test time, reaches the target as often as explicit lifting, completes the task more often, and achieves the best success rate on every TravelUAV split, with the largest gain on the unseen map.
The T1/T2 decomposition can also be applied to other goal-directed navigation policies, where it separates failures that a single success rate does not distinguish.

\section*{LLM Usage Statement}
We used Claude in an assistive role only, \ie, to edit or refine content
produced by the authors. It was used to (i) edit the language of
author-written text, (ii) help debug and refine author-written code,
(iii) suggest refinements to author-made figures and captions, and
(iv) provide reviewer-style feedback on the manuscript. No LLM or
AI tool was used to create data or annotations, or to generate research
ideas, hypotheses, claims, or conclusions. All suggestions were reviewed
and verified by the authors, who take full responsibility for all
submitted content.
{
    \small
    \bibliographystyle{ieeenat_fullname}
    \bibliography{main}
}

\clearpage
\setcounter{page}{1}
\maketitlesupplementary
\appendix

\Cref{sec:supp_depth} describes how the aerial depth model $\Phi$ of \cref{sec:explicit} is fine-tuned and evaluated.
\Cref{sec:supp_results} reports all metrics for the ablations of \cref{tab:ablation} and tests whether the failure analysis of \cref{sec:failures} depends on the reach radius.
\Cref{sec:supp_headless} discusses the headless evaluation of AeroVLA and compares \method{} with AeroVLA's reported results.

\section{Aerial Depth Model}
\label{sec:supp_depth}

Explicit lifting (\cref{sec:explicit}) needs metric depth for both cameras.
We start from the metric Depth Anything V2-Large model (DAv2-L)~\cite{yang2024depthv2}, which was fine-tuned for metric depth on driving scenes (Virtual KITTI).
Seen from the air, and especially in the downward view, its metric scale is far off (\cref{tab:depth}), so we fine-tune it on aerial images and freeze the result as $\Phi$ for all our policies.

\begin{table}[t]
\centering
\footnotesize
\setlength{\tabcolsep}{3pt}
\caption{\textbf{Depth accuracy.} Errors against simulator ground truth.
$\Phi$ is fine-tuned only on the TravelUAV training set: we split its trajectories into fine-tuning (6,090), validation (761) and test (760) trajectories, and use no episode from the benchmark's test splits.
Errors are on the validation trajectories (about 42k front/down image pairs); on the test trajectories, $\Phi$ obtains the same AbsRel.
$\delta_1$: share of pixels with $\max(\hat d/d, d/\hat d){<}1.25$. RMSE in meters.}
\label{tab:depth}
\begin{adjustbox}{max width=\linewidth}
\begin{tabular}{@{}l ccc ccc@{}}
\toprule
& \multicolumn{3}{c}{Front camera} & \multicolumn{3}{c}{Downward camera} \\
\cmidrule(lr){2-4}\cmidrule(l){5-7}
Depth model & AbsRel$\downarrow$ & $\delta_1\uparrow$ & RMSE$\downarrow$ & AbsRel$\downarrow$ & $\delta_1\uparrow$ & RMSE$\downarrow$ \\
\midrule
DAv2-L (off-the-shelf) & \underline{0.713} & \underline{0.186} & \underline{20.39} & \underline{0.873} & \underline{0.146} & \underline{13.19} \\
$\Phi$ (ours, aerial-adapted) & \textbf{0.044} & \textbf{0.977} & \textbf{2.47} & \textbf{0.035} & \textbf{0.988} & \textbf{0.85} \\
\bottomrule
\end{tabular}
\end{adjustbox}
\end{table}

\paragraph{Data.}
We use only the training set of \cref{sec:setup}.
Its 7,611 trajectories are split into fine-tuning (6,090), validation (761) and test (760) trajectories, and no episode from the benchmark's test splits is used.
The fine-tuning trajectories provide about 327k front/down image pairs.
The ground truth is the simulator's depth of each image which, like the output of $\Phi$, is the range along each pixel's viewing ray.

\paragraph{Training.}
We fine-tune the full model, encoder and DPT head, with the scale-invariant logarithmic (SiLog) loss~\cite{eigen2014depth} as used by DAv2.
Front and downward images are processed one view at a time, as at test time: each $256{\times}256$ image is resized to $518{\times}518$ with the same preprocessing as at inference, and the prediction is resized back to $256{\times}256$ for the loss.
We train for 3 epochs (61,329 steps) on four GPUs with AdamW, a batch of 16 front/down pairs, learning rates of $5{\times}10^{-5}$ for the head and $5{\times}10^{-6}$ for the encoder, weight decay 0.01, a 5\% warm-up followed by cosine decay (restarted after the first epoch), gradient clipping at 1.0, bfloat16 precision and random horizontal flips.
This one-time cost amounts to 48 GPU-hours (\cref{sec:cost}).

\paragraph{Evaluation.}
Following the official DAv2 evaluation, we compute the metrics for each image and average them over images.
For predicted depth $\hat d_p$ and true depth $d_p$ at pixel $p$, AbsRel is the mean of $|\hat d_p-d_p|/d_p$, $\delta_1$ is the share of pixels with $\max(\hat d_p/d_p,\,d_p/\hat d_p)<1.25$, and RMSE is the square root of the mean of $(\hat d_p-d_p)^2$.

\paragraph{Results.}
\Cref{tab:depth} compares $\Phi$ with the off-the-shelf model on the 761 validation trajectories.
Off-the-shelf DAv2-L misjudges metric depth badly from the air: its AbsRel is 0.713 on the front camera and 0.873 on the downward camera, and only 18.6\% and 14.6\% of pixels lie within a factor of 1.25 of the true depth.
Fine-tuning lowers AbsRel to 0.044 and 0.035 (16 and 25 times lower), raises $\delta_1$ to 0.977 and 0.988, and reduces RMSE from 20.39 to 2.47\,m on the front camera and from 13.19 to 0.85\,m on the downward camera.
For the policy, accurate depth helps most on the unseen map, where explicit lifting with $\Phi$ raises SR over off-the-shelf depth from 36.22\% to 40.08\%, and from 37.47\% to 42.69\% when combined with implicit alignment (\cref{tab:ablation_full}), whereas on the Seen split both differences are below 0.7 points.

\section{Additional Results}
\label{sec:supp_results}

\begin{table*}[t]
\centering
\footnotesize
\setlength{\tabcolsep}{3.4pt}
\caption{\textbf{Full ablation results.} All metrics for the variants of \cref{tab:ablation} on the Full, Easy and Hard episodes of each split. SR, OSR and SPL in \%, NE in meters. Bold: best in each column; underline: second best.}
\label{tab:ablation_full}
\begin{adjustbox}{max width=\linewidth}
\begin{tabular}{@{}l cccc cccc cccc@{}}
\toprule
& \multicolumn{4}{c}{Full} & \multicolumn{4}{c}{Easy} & \multicolumn{4}{c}{Hard} \\
\cmidrule(lr){2-5}\cmidrule(lr){6-9}\cmidrule(l){10-13}
Variant & NE$\downarrow$ & SR$\uparrow$ & OSR$\uparrow$ & SPL$\uparrow$ & NE$\downarrow$ & SR$\uparrow$ & OSR$\uparrow$ & SPL$\uparrow$ & NE$\downarrow$ & SR$\uparrow$ & OSR$\uparrow$ & SPL$\uparrow$ \\
\midrule
\multicolumn{13}{@{}l}{\textit{Test Seen} (1,418 episodes)} \\
AeroVLA (ECCV'26) & 75.63 & 43.86 & 52.82 & 34.37 & 47.58 & 45.34 & 56.58 & 33.34 & 110.22 & 42.05 & 48.19 & 35.64 \\
Base (no depth) & 74.29 & 44.71 & 53.17 & 34.71 & 46.66 & 45.47 & 56.70 & 33.85 & 108.36 & 43.78 & 48.82 & 35.77 \\
Base + AutoFly's depth (ICLR'26) & 76.67 & 43.16 & 51.90 & 33.56 & 50.40 & 42.53 & 54.15 & 31.87 & 109.06 & 43.94 & 49.13 & 35.64 \\
Explicit, off-the-shelf depth & 69.07 & 46.97 & 55.64 & \underline{38.43} & 46.01 & 49.17 & \textbf{60.79} & 36.99 & \textbf{97.50} & 44.25 & 49.29 & \underline{40.21} \\
Explicit & 69.24 & \underline{47.46} & 55.85 & 37.89 & 44.66 & 49.43 & 59.90 & \underline{37.75} & 99.55 & \underline{45.04} & 50.87 & 38.06 \\
Implicit & \textbf{67.55} & 47.10 & 55.57 & \textbf{39.37} & \textbf{42.98} & 49.30 & 60.28 & \textbf{38.02} & \underline{97.85} & 44.41 & 49.76 & \textbf{41.03} \\
Explicit (off-the-shelf) + implicit & \underline{68.52} & 47.39 & \underline{56.00} & 37.64 & \underline{44.18} & \underline{49.81} & 59.90 & 36.94 & 98.53 & 44.41 & \textbf{51.18} & 38.50 \\
\textbf{\method{}} (explicit + implicit) & 68.83 & \textbf{48.03} & \textbf{56.35} & 38.15 & 44.32 & \textbf{50.32} & \underline{60.66} & 37.55 & 99.05 & \textbf{45.20} & \underline{51.02} & 38.89 \\
\midrule
\multicolumn{13}{@{}l}{\textit{Test Unseen Object} (629 episodes)} \\
AeroVLA (ECCV'26) & 73.83 & 48.49 & 55.80 & 39.06 & 54.16 & 45.93 & 54.55 & 35.46 & 83.62 & 49.76 & 56.43 & 40.85 \\
Base (no depth) & 72.24 & 49.60 & 56.76 & 40.26 & 52.26 & 47.37 & 55.02 & 35.14 & 82.18 & 50.71 & 57.62 & 42.81 \\
Base + AutoFly's depth (ICLR'26) & 75.48 & 48.97 & 57.39 & 40.33 & 52.13 & 50.72 & 57.89 & 37.47 & 87.10 & 48.10 & 57.14 & 41.75 \\
Explicit, off-the-shelf depth & 69.50 & 54.21 & 58.66 & 46.11 & \textbf{46.88} & 55.02 & 59.33 & \textbf{46.14} & 80.76 & 53.81 & 58.33 & 46.10 \\
Explicit & 69.41 & 54.53 & 59.14 & 45.91 & 49.40 & 55.50 & \underline{62.20} & 42.78 & 79.37 & 54.05 & 57.62 & \underline{47.47} \\
Implicit & 68.81 & 54.37 & 58.82 & 44.80 & \underline{47.40} & \underline{56.00} & 61.24 & 39.58 & 79.46 & 53.57 & 57.62 & 47.40 \\
Explicit (off-the-shelf) + implicit & \textbf{68.27} & \underline{54.69} & \underline{59.94} & \underline{46.26} & 48.71 & 53.11 & \underline{62.20} & 44.47 & \textbf{78.00} & \underline{55.48} & \underline{58.81} & 47.15 \\
\textbf{\method{}} (explicit + implicit) & \underline{68.32} & \textbf{56.76} & \textbf{61.37} & \textbf{46.71} & 48.30 & \textbf{58.85} & \textbf{64.11} & \underline{45.03} & \underline{78.28} & \textbf{55.71} & \textbf{60.00} & \textbf{47.55} \\
\midrule
\multicolumn{13}{@{}l}{\textit{Test Unseen Map} (958 episodes)} \\
AeroVLA (ECCV'26) & 70.86 & 34.03 & 52.30 & 24.37 & 50.51 & 39.22 & 57.58 & 26.68 & 99.63 & 26.70 & 44.84 & 21.11 \\
Base (no depth) & 69.59 & 34.97 & 53.13 & 25.28 & 50.68 & 39.75 & 58.11 & 26.93 & 96.31 & 28.21 & 46.10 & 22.95 \\
Base + AutoFly's depth (ICLR'26) & 66.19 & 37.37 & 53.86 & 27.48 & 50.90 & 39.39 & 56.15 & 26.40 & 87.80 & 34.51 & 50.63 & 29.01 \\
Explicit, off-the-shelf depth & 66.98 & 36.22 & 55.01 & 27.92 & 46.11 & 41.00 & \underline{61.85} & 30.06 & 96.47 & 29.47 & 45.34 & 24.90 \\
Explicit & \underline{61.95} & \underline{40.08} & \underline{57.41} & 28.73 & \textbf{41.90} & \underline{44.03} & \textbf{62.39} & 30.04 & 90.28 & 34.51 & 50.38 & 26.89 \\
Implicit & 64.49 & 38.31 & 55.85 & \underline{30.16} & 51.04 & 39.39 & 57.75 & 29.68 & \underline{83.49} & \underline{36.78} & \underline{53.15} & \underline{30.84} \\
Explicit (off-the-shelf) + implicit & 65.13 & 37.47 & 55.01 & 29.70 & \underline{44.78} & 43.67 & 61.50 & \textbf{33.14} & 93.89 & 28.72 & 45.84 & 24.83 \\
\textbf{\method{}} (explicit + implicit) & \textbf{58.14} & \textbf{42.69} & \textbf{59.39} & \textbf{31.77} & 44.89 & \textbf{44.74} & 61.68 & \underline{31.20} & \textbf{76.85} & \textbf{39.80} & \textbf{56.17} & \textbf{32.57} \\
\bottomrule
\end{tabular}
\end{adjustbox}
\end{table*}

\begin{table}[t]
\centering
\footnotesize
\setlength{\tabcolsep}{6pt}
\caption{\textbf{Sensitivity to the reach radius} on the unseen map (958 episodes). Differences in points.}
\label{tab:radius}
\begin{tabular}{@{}c cc@{}}
\toprule
Radius & Reach: \method{} $-$ implicit & Completion: \method{} $-$ explicit \\
\midrule
10\,m & $+$8.4 & $+$3.5 \\
\textbf{15\,m} & $+$8.0 & $+$3.7 \\
20\,m & $+$7.5 & $+$3.6 \\
\bottomrule
\end{tabular}
\end{table}

\subsection{Full Ablation Results}
\label{sec:supp_ablation}

\Cref{tab:ablation_full} extends \cref{tab:ablation} to all four metrics and to the Easy and Hard episodes of each split.
The variants are those of \cref{sec:depthform,sec:expimp}.

\paragraph{Success rate.}
\method{} has the highest SR in all nine combinations of split and difficulty.
On the Seen split, where all variants with grounding lie within 1.1 points of each other, its margin over the next-best variant is at most 0.6 points; on the unseen object it reaches 2.9 points (Easy), and on the unseen map 2.6 points over all episodes and 3.0 points on Hard episodes.

\paragraph{Other metrics.}
Over all episodes, NE, OSR and SPL mostly agree with SR. The main exception is Seen, where the implicit-only model has the lowest NE and the highest SPL (67.55\,m and 39.37), although it succeeds less often than \method{}, so its successful flights are more direct.
On the unseen map, \method{} is also best in NE, OSR and SPL (58.14\,m, 59.39\% and 31.77).


\paragraph{Difficulty.}
The Hard routes of the unseen map, which are long and unfamiliar, show the largest effect of accurate depth: with implicit alignment, using $\Phi$ instead of off-the-shelf depth raises SR from 28.72\% to 39.80\%, against 43.67\% to 44.74\% on Easy routes.

\paragraph{Separate views and depth features.}
Encoding the two views separately (Base) instead of as one mosaic (AeroVLA) improves SR on every split ($+$0.9, $+$1.1 and $+$0.9 points on Seen, Unseen Object and Unseen Map).
Adding depth as features (Base + AutoFly's depth) brings no gain over Base on Seen and Unseen Object and only 2.4 points on the unseen map, whereas lifting the same depth into 3D (Explicit) improves SR over Base on every split ($+$2.8, $+$4.9 and $+$5.1 points).

\subsection{Sensitivity to the Reach Radius}
\label{sec:radius}

We hypothesize that \method{} reaches the target more often than the implicit model, thanks to explicit lifting, and completes the task more often than the explicit model, thanks to implicit alignment.
The reach radius of 15\,m is a design choice, so we test both parts of the hypothesis again with 10\,m and 20\,m (\cref{tab:radius}).
As in \cref{sec:failures}, an episode counts as reached once the UAV comes within the radius (horizontally) of the target.
At every radius, \method{} reaches the target more often than the implicit model ($p{<}0.001$, exact McNemar test) and completes the task more often than the explicit model ($p{<}0.05$, paired bootstrap pooled over five runs).
Across radii, the differences vary by less than one point (reach: $+$7.5 to $+$8.4; completion: $+$3.5 to $+$3.7).

\section{Headless Evaluation of AeroVLA}
\label{sec:supp_headless}

TravelUAV simulates each episode in Unreal Engine with AirSim, which renders the camera images and steps the physics.
Data-center GPUs such as the A100 have no display output, so the simulator renders off-screen (\emph{headless}).
We evaluate every method this way (\cref{sec:setup}); for AeroVLA, we use the checkpoint released by its authors, without retraining or modification.

\paragraph{Reported and reproduced results.}
With a physical display, AeroVLA reports an SR of 47.96\%, 56.60\% and 37.58\% on Seen, Unseen Object and Unseen Map (46.5\% over all 3,005 episodes)~\cite{xu2026aerovla}.
Evaluated headless, the released checkpoint reaches 43.86\%, 48.49\% and 34.03\% (41.7\%; \cref{tab:main}).
Running the graphics engine on compute-only GPUs, which drop frames; The collisions and sudden position jumps observed on headless GPUs are glitches of the AirSim physics engine.

\paragraph{Choice of evaluation setting.}
Display GPUs are rarely available on research clusters, and the other baselines were also evaluated headless by their authors.
More importantly, frame drops and physics glitches arise in the simulator, not in the policy.
The simulator keeps running while a policy computes its next action. In the unseen-map flight logs, this has no visible effect: on steps without a forward command, the UAV moves horizontally by a median of 5\,cm (95th percentile at most 0.6\,m) under every policy, and steps in which the UAV moves more than 5\,m beyond its command occur equally often for implicit alignment and \method{} (3.6 per 1,000 steps), despite their different decision times.
Only a common evaluation setup isolates the effect of the method, which is also what our paired tests require: every method flies the same episodes from the same start states under the same rendering.
Comparing our headless results with display results would confound the method.

\paragraph{Comparison with the reported results.}
Even against AeroVLA's display results, which favor AeroVLA, \method{} evaluated headless has the higher SR on every split: 48.03\% vs.\ 47.96\% on Seen, 56.76\% vs.\ 56.60\% on Unseen Object and 42.69\% vs.\ 37.58\% on Unseen Map (48.2\% vs.\ 46.5\% overall).
On the unseen map, which tests generalization, \method{} is also better in NE (58.14 vs.\ 67.42\,m), OSR (59.39\% vs.\ 52.92\%) and SPL (31.77 vs.\ 28.22).

\end{document}